# ABCD: Automatic Blood Cell Detection via Attention-Guided Improved YOLOX

Ahmed Endris Hasen[1*], Yang Shangming[1], Chiagoziem C. Ukwuoma[1,2,3], Biniyam Gashaw[1], Abel Zenebe Yutra[1]

[1]University of Electronics Science and Technology of China, North Jianshe Road, 610054, Chengdu, Sichuan, China
[2]College of Nuclear Technology and Automation Engineering, Chengdu University of Technology, Sichuan P.R., China
[3]Sichuan Engineering Technology Research Center for Industrial Internet Intelligent Monitoring and Application, Chengdu University of Technology, Sichuan P.R., 610059, China

**ABSTRACT**

Detection of blood cells in microscopic images has become a major focus of medical image analysis, playing a crucial role in gaining valuable insights into a patient's health. Manual blood cell checks for disease detection are known to be time-consuming, inefficient, and error prone. To address these limitations, analyzing blood cells using deep learning-based object detectors can be regarded as a feasible solution. In this study, we propose an automatic blood cell detection method (ABCD) based on an improved version of YOLOX, an object detector, for detecting various types of blood cells, including white blood cells, red blood cells, and platelets. Firstly, we introduce the Convolutional Block Attention Module (CBAM) into the network's backbone to enhance the efficiency of feature extraction. Furthermore, we introduce the Adaptively Spatial Feature Fusion (ASFF) into the network's neck, which optimizes the fusion of different features extracted from various stages of the network. Finally, to speed up the model's convergence, we substitute the Intersection over Union (IOU) loss function with the Complete Intersection over Union (CIOU) loss function. The experimental results demonstrate that the proposed method is more effective than other existing methods for BCCD dataset. Compared to the baseline algorithm, our method ABCD achieved 95.49 % mAP@0.5 and 86.89 % mAP@0.5-0.9, which are 2.8% and 23.41% higher, respectively, and increased the detection speed by 2.9%, making it highly efficient for real-time applications.

**Keywords:** Deep Learning, Object Detection, Enhanced YOLOX, Attention Mechanism, Blood Cells

## 1. Introduction

In recent years, deep learning has emerged as a research focus in medical image processing and brought significant benefits to the medical field. which can extract useful diagnostic features from medical images to handle problems in medical areas[1]. This approach has been used to analyze many types of medical imaging issues in order to extract more meaningful medical information for clinical diagnosis. When it comes to predicting and detecting abnormalities and the occurrence of diseases within the body, microscopic imaging is typically crucial. Commonly, examination results of blood cell features show the state of a person's health[2]. The appropriate cell tally assists in the diagnosis of possible diseases and related lesions, and also the early identification of the underlying pathology guide for an appropriate treatment[3]. Blood is the most crucial component of human beings. White blood cells (WBCs), red blood cells (RBCs), and platelets are the three most common categories of blood cells found in the human body. RBCs transport oxygen, whereas WBCs provide immune support, and platelets help in the progression of hemostasis. The distribution of WBC and RBC in a blood cell image is comparatively unbalanced. RBC distribution is highly dense with a wide range of sizes, but the densities of WBC and platelets are relatively sparse[3]. By recognizing and counting numerous cells, the analysis results of blood cells in microscopic images may provide critical information about a person's overall well-being.

The detection of blood cells is considered the most significant stage in monitoring our health and screening for a variety of disorders. Blood diseases such as anemia, malaria, leukemia, and others need blood cell diagnosis to get more necessary information[2]. In addition, an abnormal blood cell disorder might be a signal of a disease. As a result, the detection and counting of blood cells are crucial for the early detection and diagnosis of diseases[3]. Therefore, blood cell analysis using deep learning-based object detectors could be considered an effective tool. It also enables hematology professionals to work more quickly and precisely. In general, developing a new automated technique for blood cell detection saves a lot of time while producing more accurate results[4][5]. Traditionally, human experts were required to manually examine blood cell images in order to identify and classify the various types of blood cells present[6]. However, this process takes time, and the accuracy of the results is subjective and relies on the human analyst's expertise. In addition, in the traditional methods of blood cell analysis, there are some problems in cell detection. For instance, in areas with significant cell overlap, the accuracy of cell detection is inadequate to meet the requirements. Moreover, the variety in cell shapes and sizes is one of the major issues in blood cell detection. For example, red blood cells are spherical and similar in shape and size, but white blood cells can be irregularly shaped and vary in size. Platelets are tiny and might be difficult to identify from other types of cells. The advent of deep learning has enabled automated blood cell recognition and classification[7]. By examining massive datasets of labeled blood cell images, deep learning algorithms may learn to distinguish and categorize different types of blood cells. This method is faster than manual analysis and produces more accurate and objective findings. Recently, deep-learning approaches like YOLOv3[8] have been successfully used for blood cell detection and counting. Nevertheless, these approaches struggle to detect overlapping objects and locate bounding boxes[9][5]. Blood cells may touch, overlap, or appear in clusters in cell experiments, making it challenging for the YOLO detection approach to precisely identify each cell. To overcome these limitations and solve the low accuracy of blood cell detection, we introduce a new deep-learning-based method, which is an improved YOLOX with attention guiding. The main contributions of this article are as follows,

1. This study presents ABCD, an enhanced YOLOX for blood cell detection using full microscopic images. The CBAM attention mechanism is integrated into the backbone of the YOLOX network, and the ASFF structure is connected to the network's neck. These modifications improve the network's feature extraction and fusion capabilities, thereby increasing the model's accuracy in detecting smaller objects.
2. A new CIOU loss function is introduced in place of the standard IOU to accelerate model convergence and achieve better regression results, thereby enhancing the model's performance.
3. Lastly, various YOLO series models and Faster RCNN are employed to train and test the data and analyze their comparative results. These comparative results show that our method can perform more effectively and efficiently than any of the other existing methods.

This paper is organized as follows: In section 2, the current popular object detection methods and their applications in blood cell image detection are introduced. It also introduces the YOLOX model, which serves as the inspiration for this research. Section 3 focuses on the methods employed to enhance the YOLOX model to address the problems of blood cell image detection. In section 4, the experimental result analysis of the method is presented. Section 5 delves into a comprehensive discussion of our work. Finally, Section 6 provides a conclusive summary of our work.

## 2. Related Works

Deep neural networks like convolutional neural networks (CNNs) are inspired by biological processes[10]. Recently, deep neural networks have demonstrated remarkable performance in object detection and classification tasks [10][11]. Object detection has gained a lot of attention as an essential topic in computer vision. And, object detection methods nowadays are classified into two groups: two-stage detectors and one-stage detectors[11]. The multi-stage object detection approach consists of two stages: (1) extracting pre-selected boxes from the input image, and (2) classifying and regressing using a convolutional neural network (CNN). The R-CNN series[12]–[14] are the most well-known two-stage methods for object detection. The two-stage object recognition method achieves excellent detection accuracy, but its training processes are complicated, have a slower detection speed, and it is not convenient for real-time detection. Consequently, one-stage object detection algorithms such as SSD[15] and the YOLO series [8], [16]–[18] are developed. In the past, the detection performance of the single-stage approaches was typically lower than that of the multi-stage algorithms. With the release of YOLOv3[8] the single-stage models now exceed the multi-stage models in terms of accuracy while maintaining higher detection speed. In our work, we used the one-stage algorithm YOLOX[18] as the baseline, which is faster and more precise than others.

Nowadays, several researchers have utilized a variety of the latest two-stage and one-stage algorithms in the medical field, particularly for image analysis, with the purpose of extracting the main features of images for use in the automatic diagnosis of diseases. A literature survey, however, revealed that there are just a few studies on the detection and counting of blood cells in a blood cell image as some mentioned in table 1. In[19], Zhang et al. (2019) deploy the YOLOv3 network and image density estimation to count RBCs and WBCs in microscopic images. In the BCCD dataset, the method achieves an accuracy of 83.28%. The FED detector[4] updated the mAP to 88.33% with around 14 million parameters. In 2022, Xu et al.[9] propose TE-YOLOF to increase the precision of RBC detection. EfficientNet is used as the model's backbone for feature extraction, while Depthwise Separable Convolution is used to minimize parameters. The method achieved a magnificent mean average precision (mAP) of 91.90% in the BCCD dataset. In 2019, Alam and Islam[6] introduced a method based on YOLO with different networks to detect and counter RBCs, WBCs, and platelets. A study conducted by Liu et al.[20] in 2021 introduced an improved deep learning model called ISE-YOLO, which is based on the YOLO architecture and includes a squeeze-and-excite attention module. On the Enhanced BCCD dataset, the model had a high mAP value of 85.7% and a processing speed of 34.5 frames per second (FPS) in blood cell detection. In 2020, Kutlu et al.[21] employed the ResNet50-RCNN model to detect blood cells in two datasets, Enhanced BCCD and LISC, and achieved a mAP value of 74% accuracy. Tavakoli et al.[22] employed thresholding and SVM (Support Vector Machine) approaches in 2021 to detect white blood cells (WBCs) in three datasets, Raabin-WBC, BCCD, and LISC. Xia et al.[23] (2020) use YOLOv3, YOLOv3-SPP, and YOLOv3-tiny object detectors on the BCCD dataset. Using YOLOv3-SPP and an IOU of 0.5, they achieved a mAP value of 88.6%. The other two detectors had a mAP value of around 80%.

Attention mechanisms are a highly effective optimization structure in deep learning[24]. Different kinds of attention mechanisms are utilized in computer vision tasks. SENet (Squeeze-and-Excitation Networks)[25] is a representative example of channel domain attention. It uses global average pooling to generate per-channel statistics from feature maps, two fully connected layers to learn weighted values for each channel, and a weighted feature map serves as the network input for the next layer. Consequently, a variety of researchers provided a strategy for mixed attention that integrates spatial and channel attention. For instance, in CBAM[26], which adopts a serial approach, the channel attention module is the first to be applied to the feature map, followed by the spatial attention module. Li et al.[27] utilized channel attention with YOLOv5 to detect diseases in maize leaves, while Wang et al.[28] used a deep learning strategy that included YOLO v5 with attention mechanisms for the recognition of the invasive weed Solanum rostratum Dunal seedlings. Li et al.[29] established a new method for mixed attention and used it to segment cells in the U-Net network.

Gu and Sun (2023)[30] introduce an improved YOLOv5 model, AYOLOv5, which incorporates the attention mechanism to enhance feature extraction in regions with high cell density. They leverage the Convolutional Block Attention Module (CBAM) and transformer encoder block to increase the weight of cell-dense regions, improving the network's ability to focus on cellular information. And achieve an mAP, of 93.3%. Similarly, Rahaman et al. (2022)[31] focus on the identification and counting of blood cells, a critical task for diagnosing and treating various disorders. They propose a deep learning-based model utilizing the YOLOv5 model, achieving an average precision of 0.824. The study of avian blood cells also benefits from deep learning models, as demonstrated by Vogelbacher et al. (2024)[32]. Their approach automates the quantification of avian red and white blood cells in whole slide images, employing two deep neural network models to achieve high precision and recall.

Ge et al.[18] proposed YOLOX, a high-performance object detection model that is an upgrade to the YOLO series. And it's one of the best detection network models. The backbone of YOLOX inherits the CSPNet53 network structure of YOLOv3[8]. When it comes to the neck, YOLOX utilizes the PAFPN[33] structure to achieve multi-scale feature integration. The decoupled head of YOLOX used two convolutional blocks for classification and regression, which were then integrated for prediction, as opposed to earlier YOLO series networks, which used a single 1x1 convolution for both tasks. Because the anchor-based prior box mechanism used by the earlier YOLO series networks resulted in data redundancy, YOLOX utilized the anchor-free prior box mechanism[34], which not only requires less time but also makes the YOLO head less complicated. SimOT, a dynamic matching mechanism, is another notable improvement of YOLOX[35].

Table 1. Comparison Table of Related Works

| Reference | Year | Methods | Dataset | No. of Images | Performance |
|---|---|---|---|---|---|
| Alam and Islam et al. [6] | 2019 | VGG16-YOLO ResNet50-YOLO Inceptionv3 - YOLO MobileNet-YOLO | BCCD | 364 | mAP = 71.32% mAP = 74.37% mAP = 68.26% mAP = 52.07% |
| Zhang et al. [19] | 2019 | YOLO-Tiny YOLOv3 and Image density estimation | BCCD | 364 | mAP = 83.29% |
| Xia et al. [23] | 2020 | YOLOv3-SPP YOLOv3 | BCCD | 364 | mAP = 88.6% mAP ≈ 81.0% |
| Kutlu et al. [21] | 2020 | ResNet50-RCNN | Enhanced BCCD and LISC | 6250 | mAP = 74% |
| Shakarami et al. [4] | 2021 | FED | BCCD | 364 | mAP = 89.86% |
| Liu et al. [20] | 2021 | ISE-YOLO | Enhanced BCCD | N/A | mAP = 85.7%, 34.5 FPS |
| Tavakoli et al. [22] | 2021 | Thresholding and SVM | Raabin-WBC BCCD LISC | 14,514 257 349 | mAP = 94.65% mAP = 92.21% mAP = 94.20% |
| Xu et al. [9] | 2022 | TE-YOLOF | BCCD | 364 | mAP = 91.90% |
| Rahaman et al. [31] | 2022 | DCBC DeepL | BCCD | 366 | mAP = 0.824 |
| Gu and Sun et al. [30] | 2023 | AYOLOv5 | BCCD | 364 | mAP = 93.3% |
| Vogelbacher et al. [32] | 2024 | Deep Neural Network Models | avian blood smear samples | 1810 | F1 score = 95.3%, mAP = 90.7% |

The research motivations of this paper are as follows:
- Manually detecting blood cells in a microscopic image is time-consuming, inefficient, and errors in results.
- In the traditional methods of blood cell analysis, there are some problems. For instance, in the regions where cells overlap, the accuracy of cell detection falls short of meeting the requirements.
- Despite the successful use of deep-learning-based detection methods like YOLOv3 in detecting blood cells, these approaches encounter difficulties in distinguishing overlapping objects and locating the bounding box.

To address these issues and significantly improve the efficiency of the detection results. In this study, an improved YOLOX with an attention-guiding method for the detection of the three major categories of blood cells, white blood cells, red blood cells, and platelets is proposed.

## 3. Materials and Methodology
### 3.1. Method Overview

The overview of the proposed ABCD method is illustrated in Figure 1. First, various data augmentation techniques are applied to enhance the number of base datasets for training to increase the generalization ability and performance of the model. Then, the CBAM[26] module is added to the backbone of the network as well as the ASFF[36] structure is added to the neck of the network, to improve the network's feature extraction and fusion, resulting in improved performance in recognizing small objects. Additionally, the IOU (intersection over union) loss function is replaced with the CIOU (complete intersection over union)[37] loss function to address the overlapping area problem and accelerate the convergence of the model.

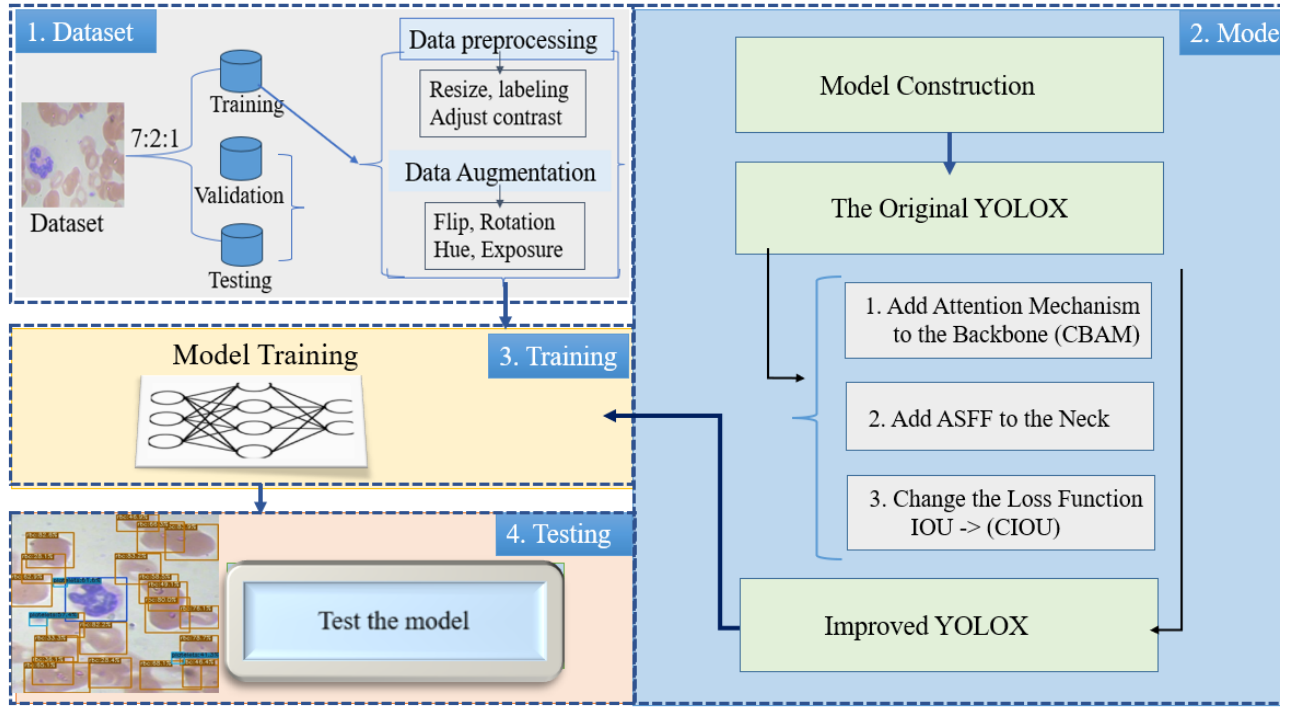

Figure 1. The Overall Architecture of The Proposed Method

In general, the proposed method is described below step by step:
- Step 1: The images from the BCCD dataset that have been augmented are loaded into a dataset.
- Step 2: In this step, we develop an improved YOLOX model, and the overall architectural modification of this model is described as follows:
  - Firstly, CBAM is added to the backbone of the network.
  - Secondly, ASFF is linked with the neck of the network.
  - Lastly, CIOU loss function is used instead of the IOU.
- Step 3: The improved YOLOX, is trained using the training data that are loaded from the datasets in Step 1. After the training phase, the proposed model is evaluated by the Val and test sets. The primary performance measure is mean average precision (mAP). Additionally, the number of parameters and FPS (frame per second) is utilized to evaluate the effectiveness of our method.
- Step 4: In this step, the visualization results of the proposed detector method have been presented.

### 3.2. Improved YOLOX

YOLOX[18] is one of the most efficient detectors widely available today, which preprocesses data using effective data augmentation techniques. It's a frame-based detector that does not use anchors to avoid the issue of unbalanced positive and negative samples that result from the anchor frame method. Also, in YOLOX decoupled heads work simultaneously on classification and regression tasks, which is significantly faster and more accurate than other detectors. Therefore, YOLOX was chosen as the baseline and made some improvements to it. The basic improvements are adding the CBAM[26] to the backbone of the network to increase the effectiveness of feature extraction in the network and connecting the ASFF[36] structure to the neck of the network to optimize feature fusion. Additionally, the IOU loss function is replaced with the CIOU loss function[37], to balance the ratios of positive and negative samples and to accelerate the convergence of the model. Based on this improvement, an improved YOLOX is proposed.

The sketch of our improved YOLOX model is depicted in Figure 2.

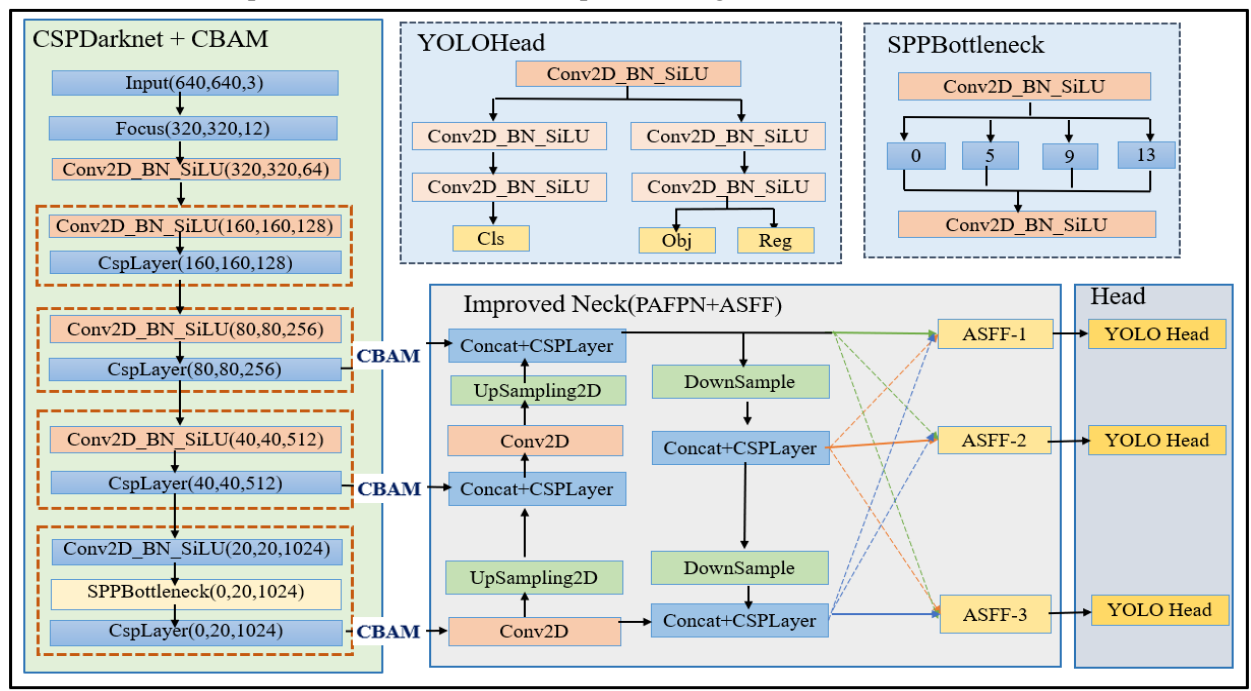

Figure 2. Improved YOLOX Architecture

The architecture in Figure 2 has three components: The backbone, Neck, and Head. The backbone adopts CSPNet53 from YOLOv3[8] as the feature extractor. It improves gradient flow and reduces computation using Cross Stage Partial (CSP) connections, enabling effective feature extraction. CSPNet53 contains 53 convolutional layers organized into different stages. Each stage consists of multiple blocks. The input to each stage is first passed through a focus layer to reduce channels. It is then fed into a series of Conv2D_BN_SiLU and CSP blocks within the stage. Each stage in the backbone consists of a Conv2D_BN_SiLU layer followed by a CSP Layer, which fuses input and output features for enhanced learning to reduce dimensions. Between the Conv2D_BN_SiLU layers there is a Cross Stage Partial (CSP) layer that fuses the input and output of the blocks to combine features. After that, there is a Spatial Pyramid Pooling (SPP) layer for spatial information fusion before the features are passed to the next stage. Additionally, CBAM[26] is integrated with the last three stage after the CSP layer. The CBAM modules generate channel and spatial attention weights, which are multiplied by the CSP features to highlight important channels and spatial regions, enhancing the feature representations learned by the backbone. The neck part connects the output features from the last three stages of the CSPNet53 backbone and passes through a conv2d layer to reduce channels followed by upsampling to match the spatial size of the second last stage. These upsampled features are concatenated with the features from the second last stage. This concatenated output acts as the input to a Cross Stage Partial (CSP) layer which fuses the multi-scale information. Similarly, the output of the first CSP layer is upsampled, concatenated with the features from the third last stage, and fed to another CSP layer. Additionally, an Adaptively Spatial Feature Fusion (ASFF) module[36] is integrated which generates spatially adaptive weighting vectors based on the features. The head part contains separate branches for classification and bounding box regression. It takes the fused features from the neck as input. The features are divided between the two branches. In the classification branch, the features are passed through the convolutional layers to predict the probabilities for object and class confidence scores. Similarly, the regression branch passes the features through the convolutional layers to predict the bounding box coordinates.

## A. Convolutional Block Attention Module (CBAM)

The main goal of the attention mechanism is to improve the representation of the region of interest by getting the neural network layer to focus more on the features that we want[24]. Recently, several attention mechanisms have been used to enhance the network's performance on object detection tasks. CBAM[26] is one of the most effective attention mechanisms. The CBAM is a minimal convolutional attention module that combines spatial and channel attention mechanisms. This combination enables CBAM to efficiently and accurately identify key features while suppressing irrelevant or unimportant data[35].

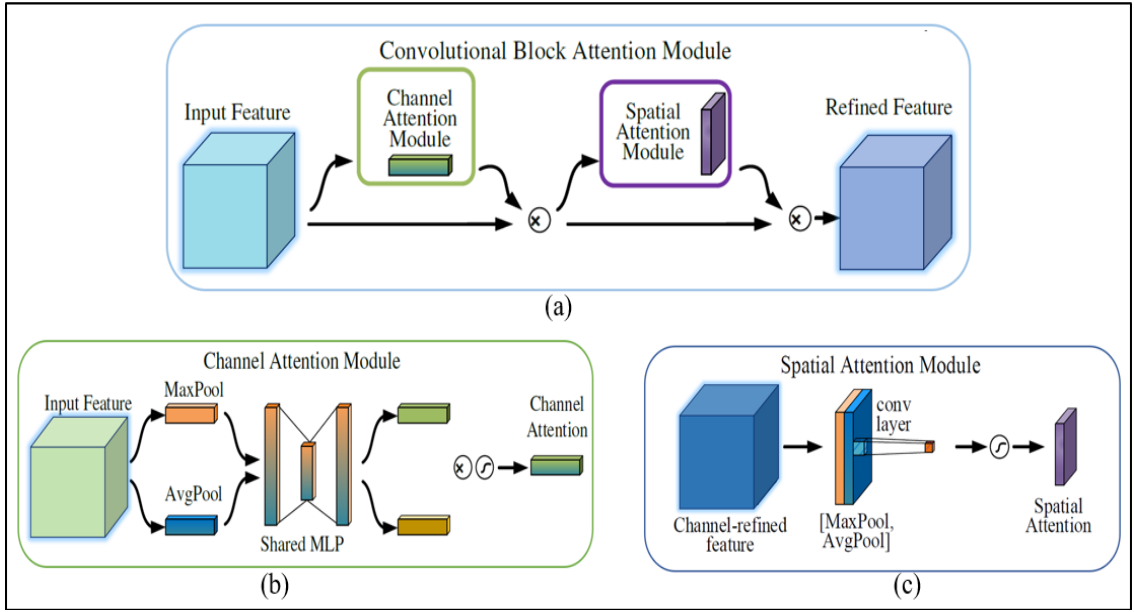

Figure 3. The Figure Illustrates the Attention Techniques[26]. (A) CBAM; (B) Channel Attention; And (C) Spatial Attention

The convolutional block attention is comprised of input, a channel attention module as shown in figure 3(b), a spatial attention module in figure 3(c), and output as demonstrated in Figure 3(a). Input features $F \in R^{C*H*W}$, undergo one-dimensional convolution using the channel attention module $M_c \in R^{C*1*1}$, after which they are multiplied by the original image. The resulting product serves as input for performing two-dimensional convolution using the spatial attention module $M_s \in R^{1*H*W}$, which is then multiplied by the original image based on Eq. (1):

$$\begin{aligned} F' &= M_c(F) \otimes F, \\ F'' &= M_s(F') \otimes F' \end{aligned} \quad (1)$$

where $\otimes$ represents element-wise multiplication. The final output is denoted by the letter $F''$. Attention values are broadcast properly during multiplication: channel attention values are broadcast along the spatial dimension and vice versa. The channel attention mechanism[38] in the CBAM module is realized by performing maximum pooling and average pooling operations on the feature map, then processing and adding the generated feature map using the shared full connection network, and finally activating it using the sigmoid function. The relevant formula is defined as follows:

$$M_c(F) = \sigma(\text{MLP}(\text{avgpool}(F)) + \text{MLP}(\text{maxpool}(F))) \\ = \sigma\left(W_1\left(W_0(F_{avg}^c)\right) + W_1\left(W_0(F_{max}^c)\right)\right), \quad (2)$$

Where $\sigma$ is the sigmoid function, $W_0 \in \mathbb{R}^{C/r \times C}$, and $W_1 \in \mathbb{R}^{C \times C/r}$. $W_0$ and $W_1$ of the MLP weights are shared by both inputs, and $W_0$ is activated using the ReLU function. The spatial attention mechanism[39] calculates the maximum and average value on the channel of each feature point and stacks them, then adjusts the number of channels of the feature map using convolution, and lastly activates it using the sigmoid function. The relevant formula is as follows:

$$M_s(F) = \sigma(f^{7 \times 7}([\text{avg pool}(F); \text{maxpool}(F)])) \\ = \sigma(f^{7 \times 7}([F_{avg}^s; F_{max}^s])), \quad (3)$$

Where $\sigma$ is the sigmoid function and $f^{7 \times 7}$ is a convolution operation with a filter size of $7 \times 7$.

### B. Adaptively Spatial Feature Fusion (ASFF)

Besides utilizing CBAM to increase the number of features obtained by the network backbone, in this paper an efficient feature fusion technique called ASFF[36], to optimize the feature fusion of the network is used. The architecture of YOLOX's neck network uses a multi-level feature pyramid, PAFPN[33]. PAFPN combining semantic and spatial information to improve object detection performance. The structure enables the neck network's output to merge location information from the shallow network with semantic data from the deep network, optimizing regression and classification outcomes. ASFF could be added into PAFPN to conduct feature fusion on the PAFPN-processed feature map, which can enhance detection performance by combining related features. The ASFF module could instruct each feature layer to identify objects that correspond to its grid size, spatially filter features on other levels, and retain just the data required for composition. This solves the issue of overlapping, indistinguishable images of varying sizes in blood cells[5].

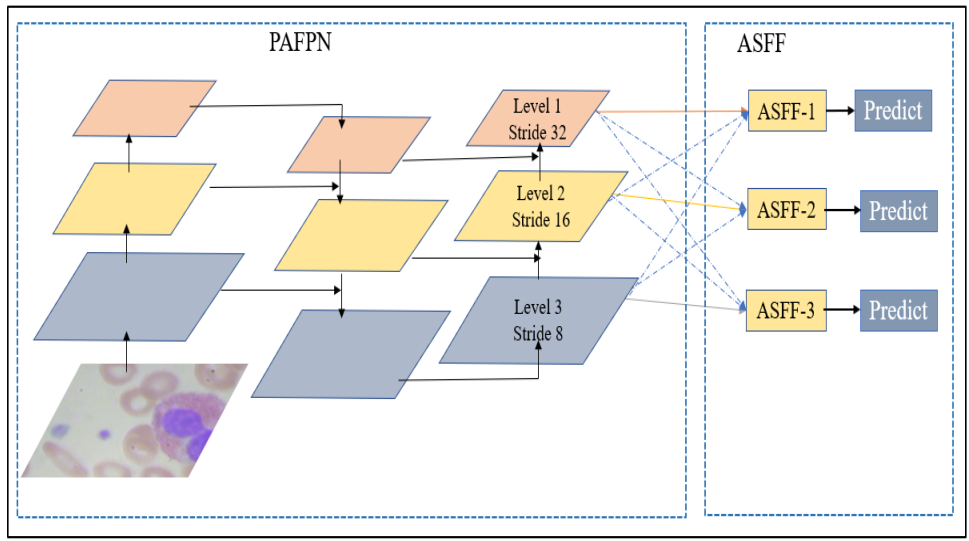

Figure 4. Structure of Connecting PAFPN With ASFF

The image will generate three feature maps through PAFPN, as demonstrated in Figure 4. They are labeled from top to bottom as follows: Level $X^1$, level $X^2$, and level $X^3$. $X^1$ has the largest receptive field and can identify large items in the image, $X^3$ has the lowest receptive field and can detect tiny objects, and $X^2$ has a midlevel of the receptive field and can detect medium-sized objects. Three feature maps are fused by ASFF using the PAFPN network output $(X^1, X^2, X^3)$ as an input. The ASFF calculating procedure consists of two parts: resizing of features and adaptive fusion. Scaling a feature map to the fused feature map's size in order to maintain the fused feature map's original size is known as feature resizing. We represent the image with $X^{1->3}$ after scaling $X^1$ to the size of $X^3$. Adaptive fusion is utilized to train three weight maps α, β, and γ in the network, which are then multiplied with the input feature maps. Using the input of ASFF3 as an example, the following calculation is carried out:

$$\text{ASFF3} = X^{1->3} \otimes \alpha^3 + X^{2->3} \otimes \beta^3 + X^{3->3} \otimes \gamma^3 \tag{4}$$

This computation enhances the features of tiny objects in the feature map, $X^3$ utilized to identify smaller objects. Similarly, the activation values of features in the feature map $X^3$ of large and medium objects are filtered out, therefore the model focuses more on identifying smaller objects. For ASFF1 and ASFF2 the same principle is utilized. Both incorporate the characteristics of their own concern from the three feature maps and reduce unnecessary information, hence enhancing the accuracy of object detection[40].

### C. Loss Function

The loss functions for bounding boxes (IOU loss[41] and GIOU loss[42]) for predicting regression have significant limitations, making it nearly impossible to maximize the overlap between the actual box and the detection box when one is inside the other. Likewise, the original approach used a binary cross-entropy loss function for the category loss and confidence degree, which is inconsistent with the positive and negative sample classification[43]. To improve accuracy and performance while significantly accelerating convergence rate, this approach utilizes CIOU loss as a regression bounding box loss function that enhances the aspect-ratio limitation method defined in Eq. (5). The consistency of aspect ratio is measured using Eq. (6), and confidence degree and category losses are calculated using PolyLoss function based on Taylor expansion approximation of focal loss[44]. This approach not only ensures that prediction boxes align better with real boxes.

$$\text{CIOU Loss} = 1 - \left(IOU - \frac{\rho^2(b, b^{gt})}{c^2} - \alpha v\right) \tag{5}$$

$$v = \frac{4}{\pi^2}\left(\arctan\frac{w^{gt}}{h^{gt}} - \arctan\frac{w^p}{h^p}\right)^2 \tag{6}$$

$$\alpha = \frac{v}{(1-IOU)+V} \tag{7}$$

The weight coefficient α, the distance between the two frames' aspect ratios v, the Euclidean distance between the two boxes' centers ρ(), and the diagonal length of the smallest circumscribed rectangle of the two c are all used in the above equations. The reason for replacing the IOU loss function with the CIOU loss function is to accelerate the convergence of the model and to address the problem of overlapping areas between the ground truth and predicted boxes, which can limit the ability to maximize the overlap. The CIOU loss improves upon IOU by incorporating the aspect ratio consistency between boxes and the distances between box centers. This allows better alignment of prediction boxes with the true boxes when one is enclosed within the other.

### 3.3. Dataset and Data Augmentation

The dataset utilized in this study is the BCCD blood cell dataset, which is an open-source and publicly available collection of annotated blood smear images that have been widely used in the development and evaluation of image analysis algorithms for blood cell detection. It is created and gathered by MIT, and openly published on the Roboflow platform. The purpose of this dataset is to support researchers in developing and evaluating algorithms that can accurately identify and count blood cells. By doing so, it aims to improve the precision of medical diagnosis and treatment. The BCCD dataset includes 364 images with 4,888 annotations, each of which has dimensions of 640x480x3, with different amounts of platelets, WBC, and RBC. Figure 5 (a) exhibits an image taken from the selected sample dataset, which consists of red blood cells, white blood cells, and platelets. In this experiment, the main challenge addressed by the improved model is the presence of a substantial number of RBCs compared to the relatively minimal quantities of WBCs and platelets. Additionally, many cells are crossing and overlapping each other within the clusters of RBCs. According to the dataset's division protocol, images in this set are partitioned into training, validation, and test sets with a ratio of 7:2:1.

As data collection and annotation require a significant amount of manual effort, an augmentation of the existing data is conducted. Augmentation consists of applying transforms to our existing images in order to generate new variants and increase the number of images in the dataset[45]. As a result, models become more accurate across a wider range of use cases and the performance will be improved by expanding the variety of learning instances accessible to the model. Based on the characteristics of the datasets, flipping, hue, rotation, and exposure, techniques are applied. Thus, the number of datasets used for training, validation, and testing became 1800, 175, and 87 respectively. Figure 5 illustrates the data augmentation methods we have used in this study.

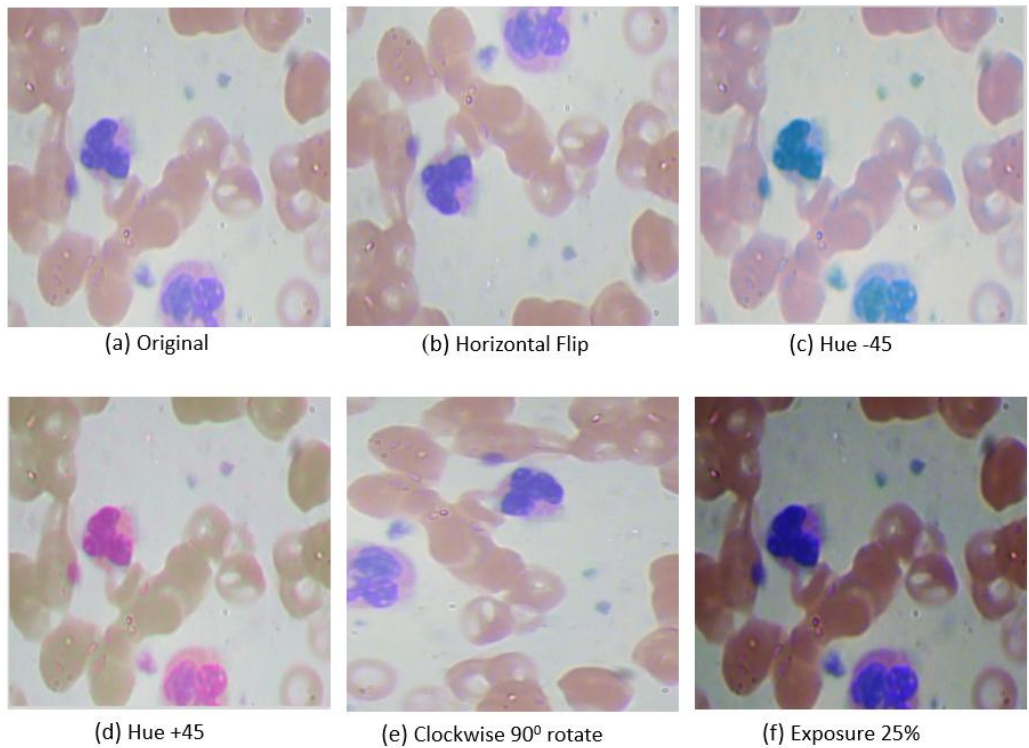

Figure 5. The Employed Augmentation Operations

### 3.4. Evaluation Metrics

In this experiment, Precision, Recall, F1-score, mAP (mean average precision), and FPS (frames per second) are utilized to assess the model's performance. Precision (P) is the proportion of actual positive samples among all expected positive samples. The recall (R) shows the proportion of positive results expected among all positive samples[46]. The recall values directly represent Sensitivity. Specificity, also known as the True Negative Rate (TNR), is a metric used to measure the accuracy of a classification model in identifying negative cases correctly, which can be obtained based on the confusion matrix. The average precision (AP) value for each class is the area under the P-R curve formed by precision and recall. The value of the mAP is the average of all class AP values. The F1 score is determined by calculating the harmonic mean of the precision and recall scores[47]. And, the formulas are defined as follows:

$$P = \frac{TP}{TP+FP} \tag{8}$$

$$R = \frac{TP}{TP+FN} \tag{9}$$

$$AP = \int_0^1 p(r)dr \tag{10}$$

$$mAP = \frac{1}{n}\sum_{i=1}^n AP_i \tag{11}$$

$$F_1\ score = 2 \times \frac{P \times R}{P+R} \tag{12}$$

Where TP is a true positive, FP is a false positive, and FN is a false negative sample.

### 3.5. Implementation Environment

The implementation tools we have used are mentioned below in Table 2.

Table 2. The Tools We Have Utilized for The Implementation.

| Hardware and Software configuration | Specific parameters |
| --- | --- |
| CPU | Intel(R) Core (TM) i7 – 10750H CPU @2.60GHz 2.59 GHz, 16 GB memory |
| GPU | NVIDIA GeForce GTX 1650 Ti |
| System | Windows 10 |
| Language | Python 3.8 |
| Python farmwork | Pytorch |
| CUDA | Version 11.4.125 |

## 4. Experimental Result and Discussion

This section focuses on the experimental results and analysis. The advantage of the proposed method is illustrated by comparing it with Faster RCNN and other YOLO series methods. Furthermore, a comprehensive ablation quantitative result analysis of each improvement is provided. The details are described below.

### 4.1. Quantitative Comparisons with Other Algorithms

The proposed method is compared with other object detection algorithms, such as Faster R-CNN [13], YOLOv3 [8], YOLOv4 [16], YOLOv5 [17], and YOLOX [18]. The results are presented in Table 3. We can find that the $mAP_{50-90}$ and $mAP_{50}$ value of our method is better than other state-of-the-art detectors, mAP50 refers to the mean Average Precision when the IoU threshold is 0.5, while mAP50-90 refers to the mAP calculated between IoU thresholds of 0.5 and 0.9. These metrics evaluate the overall performance of our methods across all classes and provide a comprehensive measure of the algorithm's ability to detect objects accurately with varying degrees of overlap. Additionally, it also has a faster detection speed than anyone of other algorithms.

Table 3. The Comparison Evaluation Results with Other Algorithms for Testing Data.

| Algorithm | P(%) | R(%) | F1(%) | mAP50(%) | mAP50-90 (%) | P(M) | FPS |
|---|---|---|---|---|---|---|---|
| Faster-RCNN | 64.42 | 86.04 | 72.00 | 78.94 | 51.94 | 137.4 | 11.2 |
| YOLOv3 | 68.90 | 86.94 | 76.44 | 90.39 | 56.94 | 61.60 | 39.4 |
| YOLOv4 | 69.20 | 91.05 | 75.00 | 91.36 | 58.30 | 64.02 | 32.7 |
| YOLOv5 | 87.50 | 88.70 | 82.45 | 92.30 | 61.82 | 7.25 | 67.2 |
| YOLOX | 90.16 | 87.23 | 88.00 | 92.69 | 63.48 | 8.94 | 69.8 |
| **ABCD** | **90.80** | **92.28** | **91.65** | **95.49** | **86.89** | **9.45** | **73.60** |

Table 3 shows the overall performance of our proposed method on the BCCD dataset, for all three classes of blood cells regardless of the specific class. It presents the average values of different evaluation metrics calculated based on the ground truths of all classes in the test data. P(M) in Table 3 refers to the number of model parameters in millions, it provides information about the model size and complexity while FPS is frame per second.

All the models (ours and those shown for comparison) were trained for a total of 50 epochs using an initial learning rate set to 0.01 with a warmup period of 5 epochs used to stabilize training at the beginning. The models were trained using the Adam optimizer with default parameter settings. We have used the freely available annotated images of BCCD datasets during the training process.

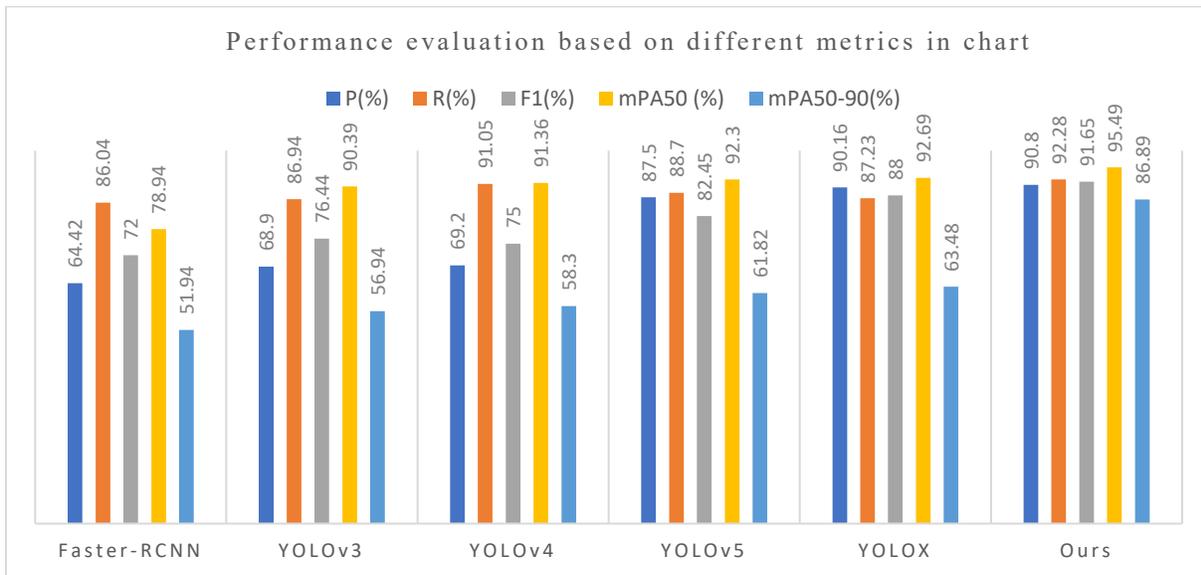

Figure 6. Comparison of Performance Evaluation Metrics Results in a Column Chart for Different Methods

Figure 6 shows a comparison of various detection models for blood cell detection, with our proposed method outperforming others in terms of precision, recall, F1-score, and mAP.

The graph in Figure 7 shows the mAP and frame per second for the proposed method and other methods for testing data. The graph demonstrated that our proposed algorithm achieved higher accuracy and speed in blood cell detection than other commonly used object detection algorithms.

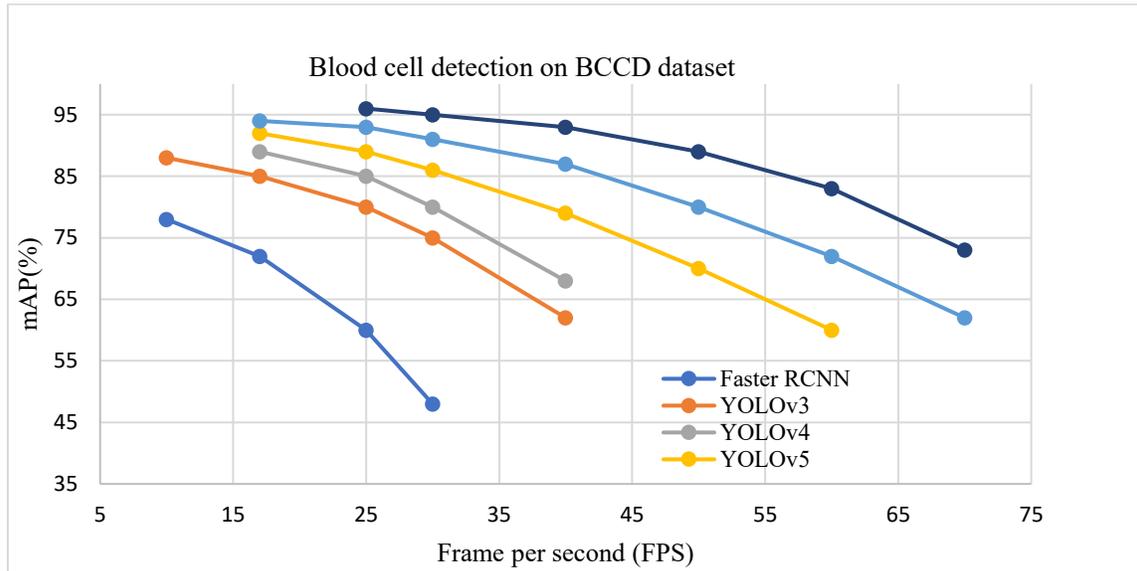

Figure 7. Accuracy And Speed Comparison of Our Method and Other Algorithms in a Graph for Testing Data

Table 4 provides the evaluation results for three categories of blood cells: Red Blood Cells (RBC), White Blood Cells (WBC), and Platelets. The evaluation metrics used are Precision (P%), Recall (R%), and Average Precision at 50% overlap (AP50).

Table 4. The Evaluation Matrix Results for The Three Categories of Blood Cells

| Blood cells | P(%) | R(%) | AP50 |
|---|---|---|---|
| RBC | 98.90 | 97.60 | 92.60 |
| WBC | 99.90 | 99.40 | 99.94 |
| Platelets | 99.30 | 93.20 | 96.49 |

The Confusion matrix below in Figure 8 evaluates the performance of the model in detecting blood cells, which includes the numerical values of classified cells (number of correctly recognized RBCs, WBCs, platelets, and number of misrecognized cells, and number of unrecognized cells for each class). This evaluation is based on the analysis of five test images comprising a total of 87 cells, with a distribution of 17 platelet cells, 57 RBCs, and 13 WBCs. The confusion matrix serves as a detailed performance metric, providing insights into the model's classification accuracy. Based on the confusion matrix, the classification results indicate that 88.24% of Platelets, 89.47% of RBCs, and 92.31% of WBCs were accurately classified.

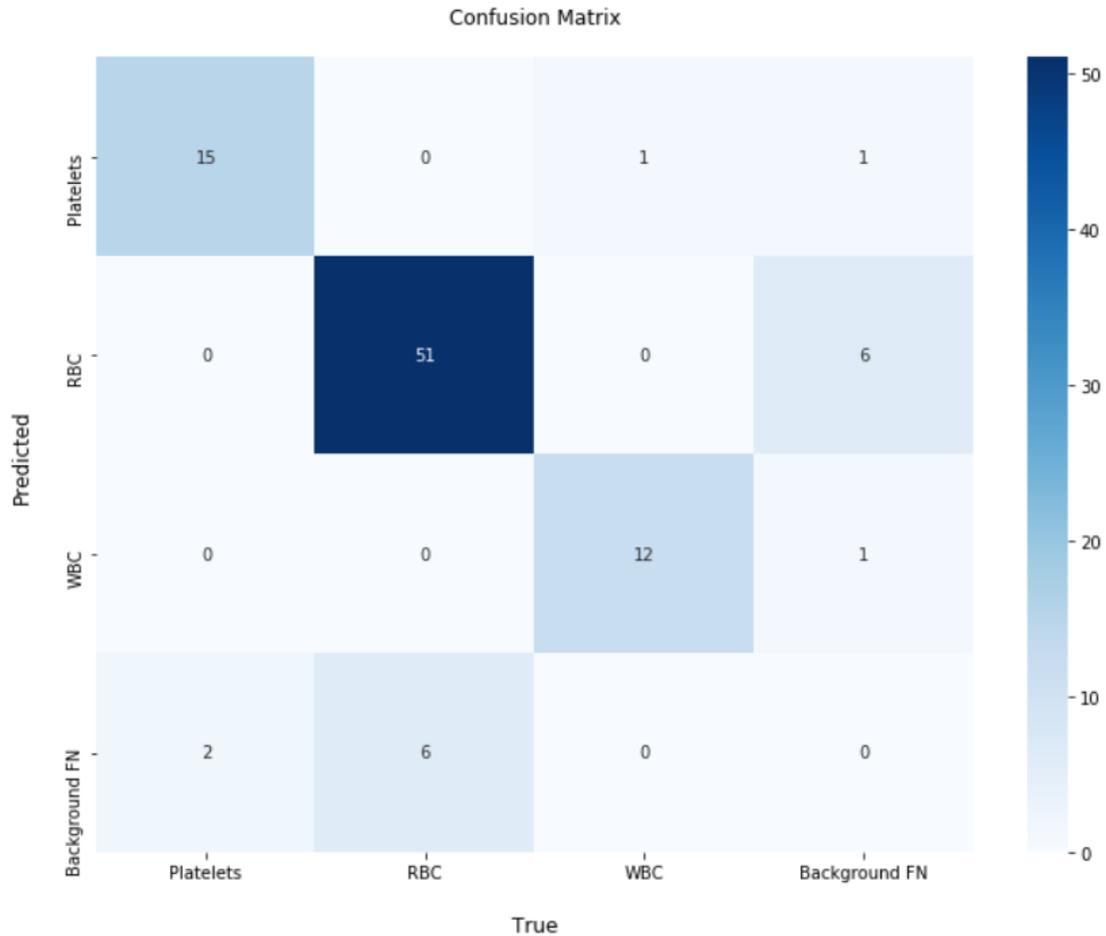

Figure 8. Confusion matrix for the numerical values of classified cells

The bar graph in Figure 9 represents the average precision of the proposed improved YOLOX blood cell detection method for the three categories of blood cells - white blood cells (WBCs), red blood cells (RBCs), and platelets. The AP50 values for each of the three classes (RBC, WBC, and platelets) are specific to each class and provide insights into the detection performance of the individual classes. In the graph, we can see that the average precision for detecting WBCs is the highest, followed by platelets and RBCs. The average precision of WBCs is around 0.99, the precision for detecting platelets is around 0.96, which is still quite good, and the precision for detecting RBCs is around 0.92. This indicates that the algorithm is more accurate in detecting WBCs than the other two types of blood cells.

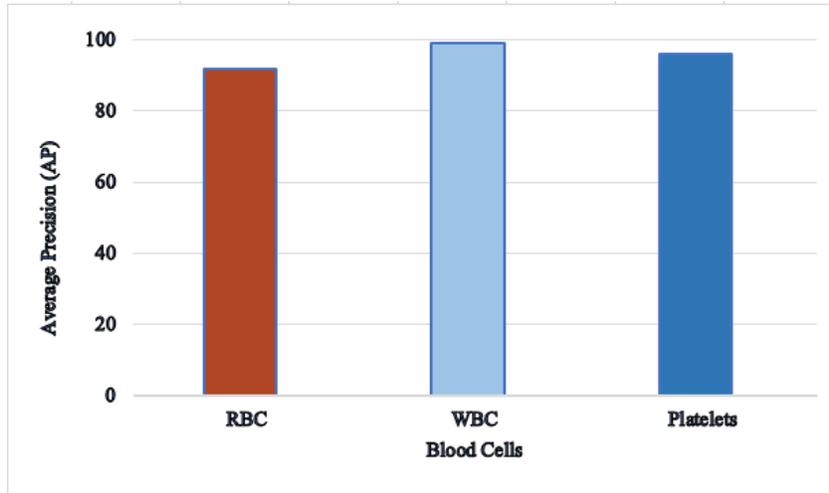

Figure 9. The Average Precision of Improved YOLOX For the Three Categories of Blood Cells

Table 5 demonstrates the performance of different methods in blood cell detection, with each method achieving varying levels of accuracy on the BCCD dataset. The proposed method in the current study shows significant improvement, achieving the highest mAP value of 95.49% among all the listed methods.

Table 5. Compares The Performance of Our Methods with Other Related Works

| Reference | Year | Methods | Dataset | Performance |
|---|---|---|---|---|
| Zhang et al. [19] | 2019 | YOLOv3 and Image density estimation | BCCD | mAP = 83.29% |
| Kutlu et al. [21] | 2020 | ResNet50-RCNN | Enhanced BCCD and LISC | mAP = 74% |
| Xia et al. [23] | 2020 | YOLOv3-SPP YOLOv3 | BCCD | mAP = 88.6% mAP ≈ 81.0% |
| Shakarami et al. [4] | 2021 | FED | BCCD | mAP = 89.86% |
| Liu et al. [20] | 2021 | ISE-YOLO | Enhanced BCCD | mAP = 85.7%, 34.5 FPS |
| Xu et al. [9] | 2022 | TE-YOLOF | BCCD | mAP = 91.90% |
| Rahaman et al. [31] | 2022 | DCBC DeepL | BCCD | Precision = 0.824 |
| Gu and Sun et al. [30] | 2023 | AYOLOv5 | BCCD | mAP = 93.3% |
| Vogelbacher et al. [32] | 2024 | Deep Neural Network Models | avian blood smear samples | F1 score = 95.3%, mAP = 90.7% |
| **ABCD** | **2024** | **Improved YOLOX** | **BCCD** | **mAP = 95.49%** |

To analyze the performance of Improved YOLOX for blood cell detection, we plot the accuracy and loss graphs per epoch for the training sets of YOLOX and the Improved YOLOX. These graphs can provide insights into how well the models are learning to detect blood cells as shown in figure 10.

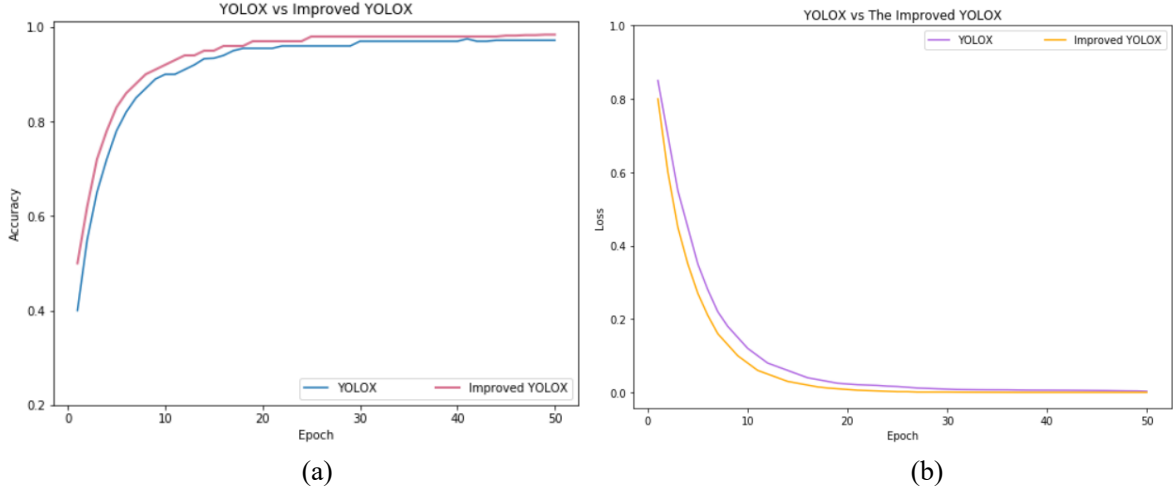

(a)             (b)

Figure 10. Shows The Accuracy and Loss Function of The Original YOLOX and for The Improved YOLOX Methods for The Training Set.

As the dataset size increases, the model's generalization ability and representation power will improve since it has access to more diverse learning samples. However, if the dataset becomes unstable due to noisy or inconsistent annotations, it can negatively impact the model. The responsiveness may reduce as stabilizing the model takes more iterations. The proposed method incorporates several techniques like data augmentation, CBAM, ASFF, and CIOU loss to make the model more robust. With a larger and well-annotated dataset, these techniques will help maintain a high level of accuracy even as the dataset size increases.

### 4.2. Ablation Study

Ablation experimental results are depicted in Table 6. The positive effects of the techniques we have applied to the original YOLOX are shown below in Table 6. The target parameters for the improved YOLOX model were determined through iterative experimentation to optimize performance. Using YOLOX as the baseline, firstly the data augmentation techniques can improve the $mAP_{50-90}$ and $mAP_{50}$ with 20.8% and 1.5% respectively. In addition, adding CBAM attention to the backbone of the network can improve the $mAP_{50-90}$ and $mAP_{50}$ with about 1.76% and 0.78% respectively. Furthermore, adding ASFF to the neck of the network improves the $mAP_{50-90}$ and $mAP_{50}$ by 0.54% and 0.24% respectively. Finally, replacing IOU loss with CIOU loss increases $mAP_{50-90}$ and $mAP_{50}$ by 0.62% and 0.32% respectively. As a result, compared with the baseline algorithm, our method achieved 95.18% $mAP_{0.5}$ and 87.2% $mAP_{0.5-0.9}$ which are 2.5% and 23.7 higher respectively.

Table 6. Ablation Experiment Results

| Methods | $AP_{50}$% | | | $mAP_{50}$(%) | $mAP_{50-90}$(%) | P(M) | (FPS) |
| --- | --- | --- | --- | --- | --- | --- | --- |
| | RBC | WBC | Platelets | | | | |
| YOLOX | 82.70 | 98.40 | 92.60 | 92.69 | 63.48 | 8.94 | 69.8 |
| +AUG. | 90.09 | 99.01 | 90.80 | 93.84(+1.16) | 84.28(+20.01) | 8.98 | 71.2 |
| +CBAM | 92.50 | 99.80 | 98.2 | 94.62(+0.78) | 86.04(+1.76) | 9.24 | 70.6 |
| +ASFF | 90.80 | 99.30 | 96.90 | 94.86(+0.24) | 86.58(+0.54) | 9.30 | 72.2 |
| +CIOU | 90.08 | 99.20 | 96.04 | 95.49(+0.63) | 86.89(+0.31) | 9.45 | 73.6 |

### 4.3. Visualization Results

The detection results of the proposed method are visualized in Figure 11. According to the visualization results, our method has the capability of detecting all of the categories efficiently. It can also recognize and identify dense RBCs and different size categories of cells accurately.

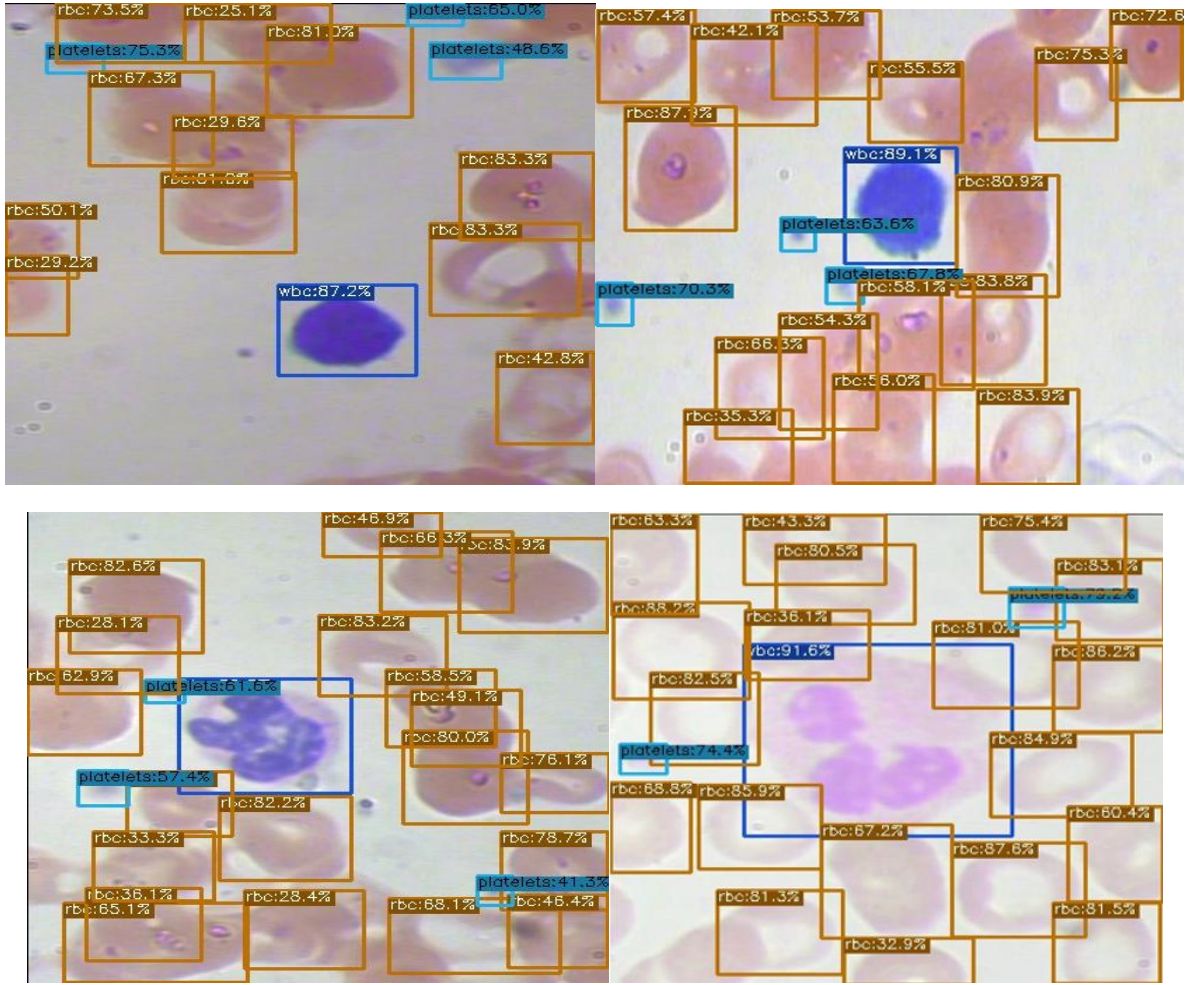

Figure 11. The Visualization Detection Result of The Proposed Method for Some Test Images

### 4.4. Discussion

Several recent works have achieved promising results applying object detection networks for blood cell analysis. However, existing methods still have limitations that our study aims to address. In this study, the proposed ABCD method for blood cell detection using an improved version of YOLOX demonstrates significant advancements over existing techniques. The experimental results presented in Table 3 demonstrate the superior performance of the proposed method compared to other algorithms, including Faster-RCNN, YOLOv3, YOLOv4, YOLOv5, and the baseline YOLOX. The proposed method achieved an impressive mAP50 of 95.49% and mAP50-90 of 86.89%, showcasing its effectiveness in accurately detecting blood cells. Additionally, the proposed method demonstrated a detection speed of 73.6 frames per second (FPS), ensuring its suitability for real-time applications.

Comparing ABCD method with related works based on Table 5, Xu et al.[9] achieved an mAP of 91.90% using TE-YOLOF on the BCCD dataset. While their results were commendable, the proposed method outperformed their approach by achieving a mAP of 95.49%, indicating its superior accuracy in blood cell detection. Zhang et al.[19] used YOLOv3 and image density estimation, achieving an mAP of 83.29% on the same BCCD dataset. In contrast, the proposed method achieved a significantly higher mAP of 95.49%, showcasing its enhanced performance. Liu et al.[20] introduced ISE-YOLO, which achieved an mAP of 85.7% and a detection speed of 34.5 FPS on the Enhanced BCCD dataset. While their approach demonstrated good accuracy, the proposed method outperformed it with a mAP of 95.49% and a higher detection speed of 73.6 FPS, highlighting its superior accuracy and efficiency. Xia et al.[23] utilized YOLOv3-SPP and YOLOv3 on the BCCD dataset, achieving mAPs of 88.6% and approximately 81.0%, respectively. The proposed method exhibited superior performance with a mAP of 95.49%, indicating its effectiveness in blood cell detection.

Gu and Sun (2023)[30] incorporated CBAM into YOLOv5, reporting an mAP of 93.3% for blood cell detection. However, our approach performs well in terms of mean average precision. Rahaman et al. (2022)[31] focused on the identification and counting of blood cells using a YOLOv5-based model. Although they reported a precision of 0.799, our method's precision on the BCCD dataset for white blood cells, red blood cells, and platelets is 99.94%, 92.6%, and 96.49%, respectively. This demonstrates our method's enhanced capability in accurately detecting and classifying individual blood cells. The two-stage approach of Vogelbacher et al.[32] achieved up to 90.7% mAP for avian cell segmentation, lower than our single-stage method which reports 95.49% and 86.89% mAP while maintaining efficiency.

The evaluation of the proposed method for the three categories of blood cells, as presented in Table 4, further reinforces its efficacy. The detection of red blood cells (RBC) achieved high precision (98.90%) recall (97.60%), and an AP50 of 92.6%, demonstrating the effectiveness of the proposed method in localizing RBCs. Regarding white blood cells (WBC), the proposed method achieved an exceptional precision of 99.90% and recall of 99.40%, indicating near-perfect detection of WBCs. The Average Precision at an IoU threshold of 0.5 (AP50) for WBCs was 99.94%, highlighting the high localization accuracy of the proposed method for WBCs. For platelets, the proposed method achieved a precision of 99.30%, and recall of 93.20%, and an AP50 of 96.49%, demonstrating the effectiveness of the proposed method in localizing platelets. The ablation experiment results in Table 6 further validate the contributions of the proposed enhancements. The addition of data augmentation techniques improved the mAP50 by 1.16%, indicating the importance of augmenting the dataset to enhance the model's generalization ability. The integration of the Convolutional Block Attention Module (CBAM) resulted in a further improvement in mAP50 (+0.78%), highlighting the effectiveness of enhancing feature extraction for smaller object detection. The Adaptively Spatial Feature Fusion (ASFF) module contributed to a slight increase in mAP50 (+0.24%), emphasizing the significance of feature fusion from different network stages. Furthermore, the replacement of the Intersection over Union (IOU) loss function with the Complete Intersection over Union (CIOU) loss function expedited the model's convergence and led to better regression results, ultimately improving the overall performance.

In general, this study presents an improved version of YOLOX for blood cell detection, surpassing the performance of existing algorithms. The proposed method achieves higher accuracy, faster detection speed, and improved efficiency, making it a promising approach for real-time blood cell analysis.

## 5. Conclusion

In this paper, we proposed an improved deep learning algorithm for blood cell detection that would significantly enhance performance over existing methods. A variety of improvements have been made to the YOLOX model. To begin, the proposed method utilizes several data augmentation techniques to enhance the number of base datasets for training to optimize the generalization ability of the model. Moreover, CBAM and ASFF were utilized to increase the model's performance in detecting small targets by enhancing the features in the backbone and neck networks. Furthermore, to solve the issue of excessive foreground-background class imbalance, we replace the IOU loss with the CIOU loss. Experiment and analysis results demonstrate that the proposed method is more efficient than other existing methods for detecting blood cell images. Finally, we expect that our detector will be used to assist medical staff and encourage the advancement of medical progress.

In our proposed method, we recognize certain limitations. One of the limitations is the relatively small size of the dataset used for blood cell detection, which may affect the generalizability of the model. A more comprehensive dataset, encompassing a wider variety of blood cell types and disease conditions, could potentially enhance performance. Additionally, the integration of modules such as CBAM and ASFF contributes to improved detection accuracy but also increases the model's complexity and computational requirements compared to the baseline YOLOX method. Exploring strategies to optimize model efficiency while preserving accuracy remains an important consideration. Another challenge involves the detection of red blood cells, which are often difficult to distinguish due to their dense distribution and overlapping characteristics. While attention mechanisms were employed to improve RBC detection and counting, further refinement may be needed to match the performance achieved with other cell types. Overall, this study demonstrates the potential of automating blood cell detection using the BCCD dataset, offering valuable support in reducing clinical workload and minimizing human error in practice.

## Reference

[1] Y. Zhang, J. M. Gorriz, and Z. Dong, "Deep learning in medical image analysis," *J. Imaging*, vol. 7, no. 4, p. NA, 2021, doi: 10.3390/jimaging7040074.

[2] G. Drałus, D. Mazur, and A. Czmil, "Automatic detection and counting of blood cells in smear images using retinanet," *Entropy*, vol. 23, no. 11, 2021, doi: 10.3390/e23111522.

[3] N. M. Deshpande, S. Gite, and R. Aluvalu, "A review of microscopic analysis of blood cells for disease detection with AI perspective," *PeerJ Comput. Sci.*, vol. 7, pp. 1–27, 2021, doi: 10.7717/peerj-cs.460.

[4] A. Shakarami, M. B. Menhaj, A. Mahdavi-Hormat, and H. Tarrah, "A fast and yet efficient YOLOv3 for blood cell detection," *Biomed. Signal Process. Control*, vol. 66, no. January, p. 102495, 2021, doi: 10.1016/j.bspc.2021.102495.

[5] E. Gavas and K. Olpadkar, "Deep CNNs for Peripheral Blood Cell Classification," pp. 1–20, 2021, [Online]. Available: http://arxiv.org/abs/2110.09508.

[6] M. M. Alam and M. T. Islam, "Machine learning approach of automatic identification and counting of blood cells," *Healthc. Technol. Lett.*, vol. 6, no. 4, pp. 103–108, 2019, doi: 10.1049/htl.2018.5098.

[7] D. Wang, M. Hwang, W. C. Jiang, K. Ding, H. C. Chang, and K. S. Hwang, "A deep learning method for counting white blood cells in bone marrow images," *BMC Bioinformatics*, vol. 22, pp. 1–13, 2021, doi: 10.1186/s12859-021-04003-z.

[8] J. Redmon and A. Farhadi, "YOLOv3: An Incremental Improvement," 2018, [Online]. Available: http://arxiv.org/abs/1804.02767.

[9] F. Xu, X. Li, H. Yang, Y. Wang, and W. Xiang, "TE-YOLOF: Tiny and efficient YOLOF for blood cell detection," *Biomed. Signal Process. Control*, vol. 73, 2022, doi: 10.1016/j.bspc.2021.103416.

[10] L. Alzubaidi *et al.*, *Review of deep learning: concepts, CNN architectures, challenges, applications, future directions*, vol. 8, no. 1. Springer International Publishing, 2021.

[11] F. Laakom, K. Chumachenko, J. Raitoharju, A. Iosifidis, and M. Gabbouj, "Learning to ignore: rethinking attention in CNNs," pp. 1–7, 2021, [Online]. Available: http://arxiv.org/abs/2111.05684.

[12] R. Girshick, "Fast R-CNN," *Proc. IEEE Int. Conf. Comput. Vis.*, vol. 2015 Inter, pp. 1440–1448, 2015, doi: 10.1109/ICCV.2015.169.

[13] S. Ren, K. He, R. Girshick, and J. Sun, "Faster R-CNN: Towards Real-Time Object Detection with Region Proposal Networks," *IEEE Trans. Pattern Anal. Mach. Intell.*, vol. 39, no. 6, pp. 1137–1149, 2017, doi: 10.1109/TPAMI.2016.2577031.

[14] X. Bi, J. Hu, B. Xiao, W. Li, and X. Gao, "IEMask R-CNN: Information-enhanced Mask R-CNN," *IEEE Trans. Big Data*, vol. 14, no. 8, 2022, doi: 10.1109/TBDATA.2022.3187413.

[15] Hao Zhang, X. gong Hong, and L. Zhu, "Detecting Small Objects in Thermal Images Using Single-Shot Detector," *Autom. Control Comput. Sci.*, vol. 55, no. 2, pp. 202–211, 2021, doi: 10.3103/S0146411621020097.

[16] A. Bochkovskiy, C. Y. Wang, and H. Y. M. Liao, "YOLOv4: Optimal Speed and Accuracy of Object Detection," *arXiv*, 2020.

[17] J. Zhao, Y. Cheng, and X. Ma, "A real time intelligent detection and counting method of cells based on YOLOv5," *2022 IEEE Int. Conf. Electr. Eng. Big Data Algorithms, EEBDA 2022*, pp. 675–679, 2022, doi: 10.1109/EEBDA53927.2022.9744842.

[18] Z. Ge, S. Liu, F. Wang, Z. Li, and J. Sun, "YOLOX: Exceeding YOLO Series in 2021," pp. 1–7, 2021, [Online]. Available: http://arxiv.org/abs/2107.08430.

[19] D. Zhang, P. Zhang, and L. Wang, "Cell counting algorithm based on YOLOv3 and image density estimation," *2019 IEEE 4th Int. Conf. Signal Image Process. ICSIP 2019*, pp. 920–924, 2019, doi: 10.1109/SIPROCESS.2019.8868603.

[20] C. Liu, D. Li, and P. Huang, "ISE-YOLO: Improved Squeeze-and-Excitation Attention Module based YOLO for Blood Cells Detection," *Proc. - 2021 IEEE Int. Conf. Big Data, Big Data 2021*, pp. 3911–3916, 2021, doi: 10.1109/BigData52589.2021.9672069.

[21] H. Kutlu, E. Avci, and F. Özyurt, "White blood cells detection and classification based on regional convolutional neural networks," *Med. Hypotheses*, vol. 135, no. October 2019, p. 109472, 2020, doi: 10.1016/j.mehy.2019.109472.

[22] S. Tavakoli, A. Ghaffari, Z. M. Kouzehkanan, and R. Hosseini, "New segmentation and feature extraction algorithm for classification of white blood cells in peripheral smear images," *Sci. Rep.*, vol. 11, no. 1, pp. 1–13, 2021, doi: 10.1038/s41598-021-98599-0.

[23] T. Xia, Y. Q. Fu, N. Jin, P. Chazot, P. Angelov, and R. Jiang, "AI-enabled Microscopic Blood Analysis for Microfluidic COVID-19 Hematology," *Proc. - 2020 5th Int. Conf. Comput. Intell. Appl. ICCIA 2020*, pp. 98–102, 2020, doi: 10.1109/ICCIA49625.2020.00026.

[24] M. H. Guo *et al.*, "Attention mechanisms in computer vision: A survey," *Comput. Vis. Media*, vol. 8, no. 3, pp. 331–368, 2022, doi: 10.1007/s41095-022-0271-y.

[25] J. Hu, "Squeeze-and-Excitation_Networks_CVPR_2018_paper.pdf," *Cvpr*, pp. 7132–7141, 2018, [Online]. Available: http://openaccess.thecvf.com/content_cvpr_2018/html/Hu_Squeeze-and-Excitation_Networks_CVPR_2018_paper.html.

[26] S. Woo, J. Park, J. Y. Lee, and I. S. Kweon, "CBAM: Convolutional block attention module," *Lect. Notes Comput. Sci. (including Subser. Lect. Notes Artif. Intell. Lect. Notes Bioinformatics)*, vol. 11211 LNCS, pp. 3–19, 2018, doi: 10.1007/978-3-030-01234-2_1.

[27]   Y. Li, S. Sun, C. Zhang, G. Yang, and Q. Ye, "One-Stage Disease Detection Method for Maize Leaf Based on Multi-Scale Feature Fusion," *Appl. Sci.*, vol. 12, no. 16, 2022, doi: 10.3390/app12167960.

[28]   Q. Wang, M. Cheng, S. Huang, Z. Cai, J. Zhang, and H. Yuan, "A deep learning approach incorporating YOLO v5 and attention mechanisms for field real-time detection of the invasive weed Solanum rostrutum Dunal seedlings," *Comput. Electron. Agric.*, vol. 199, no. January, p. 107194, 2022, doi: 10.1016/j.compag.2022.107194.

[29]   G. Li, C. Sun, C. Xu, Y. Zheng, and K. Wang, "Cervical Cell Segmentation Method Based on Global Dependency and Local Attention," *Appl. Sci.*, vol. 12, no. 15, 2022, doi: 10.3390/app12157742.

[30]   W. Gu and K. Sun, "AYOLOv5: Improved YOLOv5 based on attention mechanism for blood cell detection," *Biomed. Signal Process. Control*, vol. 88, no. April 2023, 2024, doi: 10.1016/j.bspc.2023.105034.

[31]   M. A. Rahaman, M. M. Ali, M. N. Hossen, M. Nayer, K. Ahmed, and F. M. Bui, "DCBC_DeepL: Detection and Counting of Blood Cells Employing Deep Learning and YOLOv5 Model," *Commun. Comput. Inf. Sci.*, vol. 1673 CCIS, pp. 203–214, 2022, doi: 10.1007/978-3-031-21385-4_18.

[32]   M. Vogelbacher *et al.*, "Identifying and Counting Avian Blood Cells in Whole Slide Images via Deep Learning," pp. 48–66, 2024.

[33]   S. Liu, L. Qi, H. Qin, J. Shi, and J. Jia, "Path Aggregation Network for Instance Segmentation," *Proc. IEEE Comput. Soc. Conf. Comput. Vis. Pattern Recognit.*, pp. 8759–8768, 2018, doi: 10.1109/CVPR.2018.00913.

[34]   Z. Tian, C. Shen, H. Chen, and T. He, "FCOS: A Simple and Strong Anchor-Free Object Detector," *IEEE Trans. Pattern Anal. Mach. Intell.*, vol. 44, no. 4, pp. 1922–1933, 2022, doi: 10.1109/TPAMI.2020.3032166.

[35]   M. Yu, Y. L. B, W. Shi, and Y. Xu, "An Improved YOLOX for Detection," pp. 556–567, 2022.

[36]   S. Liu, D. Huang, and Y. Wang, "Learning Spatial Fusion for Single-Shot Object Detection," 2019, [Online]. Available: http://arxiv.org/abs/1911.09516.

[37]   Z. Zheng, P. Wang, W. Liu, J. Li, R. Ye, and D. Ren, "Distance-IoU loss: Faster and better learning for bounding box regression," *AAAI 2020 - 34th AAAI Conf. Artif. Intell.*, no. 2, pp. 12993–13000, 2020, doi: 10.1609/aaai.v34i07.6999.

[38]   A. A. Bastidas and H. Tang, "Channel attention networks," *IEEE Comput. Soc. Conf. Comput. Vis. Pattern Recognit. Work.*, vol. 2019-June, pp. 881–888, 2019, doi: 10.1109/CVPRW.2019.00117.

[39]   T. Liu *et al.*, "Spatial Channel Attention for Deep Convolutional Neural Networks," *Mathematics*, vol. 10, no. 10, pp. 1–10, 2022, doi: 10.3390/math10101750.

[40]   L. Yang, G. Yuan, H. Zhou, H. Liu, J. Chen, and H. Wu, "RS-YOLOX: A High-Precision Detector for Object Detection in Satellite Remote Sensing Images," *Appl. Sci.*, vol. 12, no. 17, 2022, doi: 10.3390/app12178707.

[41]   D. Zhou *et al.*, "IoU Loss for 2D/3D Object Detection," *Proc. - 2019 Int. Conf. 3D Vision, 3DV 2019*, pp. 85–94, 2019, doi: 10.1109/3DV.2019.00019.

[42]   H. Rezatofighi, N. Tsoi, J. Gwak, A. Sadeghian, I. Reid, and S. Savarese, "Generalized intersection over union: A metric and a loss for bounding box regression," *Proc. IEEE Comput. Soc. Conf. Comput. Vis. Pattern Recognit.*, vol. 2019-June, pp. 658–666, 2019, doi: 10.1109/CVPR.2019.00075.

[43]   H. Zhang, Y. Wang, F. Dayoub, and N. Sünderhauf, "VarifocalNet: An IoU-aware Dense Object Detector," *Proc. IEEE Comput. Soc. Conf. Comput. Vis. Pattern Recognit.*, pp. 8510–8519, 2021, doi: 10.1109/CVPR46437.2021.00841.

[44]   Z. Leng *et al.*, "PolyLoss: A Polynomial Expansion Perspective of Classification Loss Functions," pp. 1–16, 2022, [Online]. Available: http://arxiv.org/abs/2204.12511.

[45]   P. Kaur, B. S. Khehra, and E. B. S. Mavi, "Data Augmentation for Object Detection: A Review," *Midwest Symp. Circuits Syst.*, vol. 2021-Augus, pp. 537–543, 2021, doi: 10.1109/MWSCAS47672.2021.9531849.

[46] A. E. Z. J. D. M. A. J. M. N. Kewyu, "Towards Automatic Ethiopian Endemic Animals Detection on Android Using Deep Learning," in *IEEE Access*, 2021, pp. 463--468.

[47] A. Endris and J. Mohammed, "Efficient Face Mask Detection Method Using YOLOX : An Approach to Reduce Coronavirus Spread," pp. 568–573, 2022.